\documentclass[journal]{IEEEtran}
\usepackage{cite}

\ifCLASSINFOpdf
\usepackage[pdftex]{graphicx}
\usepackage[table,xcdraw]{xcolor}
\else
\fi
\usepackage{amsmath}
\usepackage{amsfonts}
\usepackage{mathrsfs}
\usepackage{braket}
\newcommand{\Jaccard}{\mathcal{J}}
\newcommand{\Precision}{{\rm Pr}}
\newcommand{\Recall}{{\rm Re}}
\newcommand{\Specificity}{{\rm Spec}}
\newcommand{\Accuracy}{{\rm Acc}}

\ifCLASSOPTIONcompsoc
  \usepackage[caption=false,font=normalsize,labelfont=sf,textfont=sf]{subfig}
\else
  \usepackage[caption=false,font=footnotesize]{subfig}
\fi
\begin{document}
%
\title{Detecting Clouds in Multispectral Satellite Images Using Quantum-Kernel Support Vector Machines}
%
%
%

\author{Artur~Miroszewski,
        Jakub~Mielczarek,
        Grzegorz~Czelusta,
        Filip~Szczepanek,
        Bartosz~Grabowski,
        Bertrand~Le~Saux,~\IEEEmembership{Senior Member,~IEEE,}
        Jakub~Nalepa,~\IEEEmembership{Senior Member,~IEEE}
\thanks{A. Miroszewski, J. Mielczarek, G. Czelusta, and F. Szczepanek are with the Institute of Theoretical Physics, Jagiellonian University, {\L}ojasiewicza 11, 30-348 Cracow, Poland, e-mail: \{artur.miroszewski, jakub.mielczarek\}@uj.edu.pl, grzegorz.czelusta@doctoral.uj.edu.pl, filip.szczepanek@student.uj.edu.pl.}
\thanks{B. Grabowski is with KP Labs, Konarskiego 18C, 44-100 Gliwice, Poland, e-mail: Bertrand.Le.Saux@esa.int.}
\thanks{B. Le Saux is with European Space Agency, Largo Galileo Galilei 1, 00044 Frascati, Italy, e-mail: Bertrand.Le.Saux@esa.int.}
\thanks{J. Nalepa is with KP Labs, Konarskiego 18C, 44-100 Gliwice, Poland, and with the Department of Algorithmics and Software, Silesian University of Technology, Akademicka 16, 44-100 Gliwice, Poland, e-mail: jnalepa@ieee.org.}
\thanks{Corresponding authors: A. Miroszewski, J. Mielczarek, J. Nalepa}
\thanks{Manuscript received April 19, 2005; revised August 26, 2015.}        
}

%
%

\markboth{Journal of \LaTeX\ Class Files,~Vol.~14, No.~8, August~2015}%
{Shell \MakeLowercase{\textit{et al.}}: Bare Demo of IEEEtran.cls for IEEE Journals}
%



\maketitle

\begin{abstract}
Support vector machines (SVMs) are a well-established 
classifier effectively deployed in an array of classification tasks. In this work, we
consider extending classical SVMs with quantum kernels 
and applying them to satellite data analysis. 
The design and implementation of SVMs with quantum 
kernels (hybrid SVMs) are presented. Here, the pixels are mapped to the Hilbert space using a family of parameterized quantum feature maps (related to quantum kernels). The parameters are optimized to maximize the kernel target alignment. The quantum kernels have been selected such that they 
enabled analysis of numerous relevant properties while 
being able to simulate them with classical computers on 
a real-life large-scale dataset. Specifically, we approach the problem of cloud detection in the multispectral satellite imagery, which is one of the pivotal steps in both on-the-ground  and on-board satellite image analysis processing chains. 
The experiments performed over the benchmark Landsat-8 
multispectral dataset revealed that the simulated hybrid 
SVM successfully classifies satellite images with accuracy
comparable to the classical SVM with the RBF kernel for 
large datasets. Interestingly, for large datasets, the 
high accuracy was also observed for the simple quantum 
kernels, lacking quantum entanglement.
\end{abstract}


%
\IEEEpeerreviewmaketitle

\section{Introduction}
%
%
%
%
\IEEEPARstart{S}{atellite} imaging plays an increasingly
important role in various aspects of human activity. The 
spectrum of applications ranges from cartographic
purposes~\cite{Copernicus-Land,castillo2021semi} through 
meteorology~\cite{weather-forecast-satellites}, ecology, 
and agronomy~\cite{Nalepa2022icip} to security and urban monitoring~\cite{audebert2017deep}. Consequently, dozens of terabytes of raw imaging data are generated daily from satellite constellations, such as those built within the European Copernicus Programme. The large volume of multi- or hyperspectral images, which capture the detailed characteristics of the scanned materials, makes them difficult to transfer, store, and ultimately analyze. Therefore, their reduction through the extraction of useful information is a critical issue in real-world applications. An important step in the data analysis chain of optical satellite data is the identification of clouds. The interest is two-fold: on the one hand, such cloudy regions can be removed from further processing, as the objects of interest are likely to be obscured. On the other hand, efficient detection of cloud cover on the Earth's surface is important in meteorological and climate research~\cite{9554170}. 
Since the reduction is performed on a huge amount of raw data, the efficiency of this process is a key factor in practice. Therefore, it is reasonable to search for new methods to analyze such huge datasets, improving image data classification into clear and cloudy areas. 

\subsection{Contribution}

In this paper, we investigate the possibility of using 
quantum machine learning algorithms \cite{Biamonte_2017} 
in this context. Even though it is still in its infancy, 
the potential of quantum computations might be a game 
changer for such applications (see Refs. 
\cite{9531639,Henderson,Sebastianelli,Gupta2023}). 

Specifically, we compare the classification performance 
of a classical Support Vector Machine (SVM) and its 
quantum extensions employing quantum 
kernels\footnote{Note that Quantum Support 
Vector Machine (QSVM)\cite{Rebentrost_2014} algorithm 
has theoretically been proven to exhibit exponential 
speedup over the classical SVM. However, the full
application of the algorithm requires many
qubits, being of the order of the size of the 
training set. This is not the case for the hybrid 
SVM approach (with the quantum kernel) considered here.}. 
There are theoretical arguments \cite{havlivcek2019supervised,goldberg2017complexity,
demarie2018classical} 
that some relevant quantum kernels are hard to 
evaluate on a classical computer. Therefore, if 
they provide an advantage in classification accuracy, 
this would advocate a strong use case for quantum 
computing methods. In this article, a family of 
quantum kernels has been selected such that both 
the role of quantum entanglement can be investigated
and the quantum kernels can be studied for complex
datasets. 

Additionally, to get a deeper understanding of 
the quantum kernel methods and show their usefulness 
in practice, it is pivotal to focus on widely 
adopted image data corresponding to real use cases. 
Thus, we tackle the cloud detection task in satellite 
image data, which is one of the most important 
processing steps for such imagery. Our experimental
study was performed over the benchmark multispectral 
image data acquired by the Landsat-8 satellite revealed 
that SVMs with quantum kernels offer a classification 
accuracy at least comparable to classic RBF kernel SVMs. 

\subsection{Structure of the Paper}

This paper is structured as follows. In Section~\ref{sec:SVM}, we discuss the theory behind SVMs, quantum kernel methods and Kernel Target alignment. The proposed hybrid SVMs are presented in Section~\ref{sec:methods}. In Section~\ref{sec:validation}, we report and discuss the results of our experimental study. Finally, Section~\ref{sec:conclusions} concludes the paper and highlights the future activities which may emerge from the research presented here.

\section{Theory}

This section provides a gentle introduction to SVMs (Section~\ref{sec:SVM}). Additionally, we present the background behind the quantum kernel methods (Section~\ref{sec:QKM}) and Kernel Target alignment (Section~\ref{sec:KTA})---these concepts are exploited in our hybrid SVMs for multispectral satellite data analysis.

\subsection{Support Vector Machines}\label{sec:SVM}

In binary classification, we assign one of two labels, conventionally $\{-1, 1\}$, to each datum in a set based on its features. Considering the data in terms of points occupying a feature space, the problem can be thought of as dividing the said feature space so that each of its two parts contains only one class of data points. There are a plethora of supervised machine learning classifiers for this task, with SVMs being one of the most widely-used and well-established in the field, already exploited in an array of pattern recognition and classification tasks~\cite{nalepa2019selecting}. In SVMs, based on training data, a hyperplane is found, defined by its normal vector $w$ and offset $b$, such that for any training datum $x_i$ and its label $y_i$, we have:
\begin{equation} \label{eq:svm-hard-margin}
y_i (w \cdot x_i - b) \ge 1.
\end{equation}
In order to decrease the risk of new data being misclassified, one aims as well to maximize the margin $2 / ||w||^2$, that is, the distance between the two-class vectors~\ref{eq:svm-hard-margin}. Having found a separating hyperplane, it can be observed that it is defined by a (usually a very small) subset of training vectors, called the \textit{support vectors}, satisfying either $w x - b = 1$ or $w x - b = -1$.

The above formulation leads to a \textit{hard-margin} SVM, disallowing for any points to fall inside the margin. This makes it impossible to train the classifier on linearly non-separable data. However, a \textit{soft-margin} SVM can be introduced by allowing each datum $x_i$ to deviate by $\xi_i$ from satisfying the conditions in Equation~\ref{eq:svm-hard-margin}, obtaining a new set of conditions:
\begin{equation}
y_i (w x_i + b) \ge 1 - \xi_i.
\end{equation}
For an $N$-element training set, the optimization problem at which one arrives is in its dual form given by
\begin{equation} \label{eq:svm-dual-problem}
\begin{aligned}
\text{maximize:} \quad & \sum_{i=1}^N \alpha_i - \frac{1}{2} \sum_{i, j = 1}^N \alpha_i \alpha_j y_i y_j \langle x_i, x_j \rangle, \\
\text{subject to:} \quad & \sum_i y_i \alpha_i = 0, \; 0 \le \alpha_i \le C,
\end{aligned}
\end{equation}
where $C \ge 0$ is the regularization parameter that specifies the impact of values $\xi_i$ on the cost function.
Then, the decision function for classifying new data $x$ takes the form of
\begin{equation} \label{eq:svm-decision-function}
    f(x) = \text{sgn}\left(\sum_{i = 0}^N y_i \alpha_i \langle x, x_i \rangle + b\right).
\end{equation}

Observe that both training and test phases do not depend directly on the data points $x_i$, but on the overlap between points calculated with inner product. If we introduced a different similarity measure between points, the procedure would not change.
Therefore, SVM lends itself to the use of the \textit{kernel trick}.
With a non-linear transformation $\phi$ chosen, any potential occurrences of $\langle \phi(x_i), \phi(x_j) \rangle$, the inner product of two data points in a higher-dimensional space, can instead be replaced with the value of a kernel function $k(x_i, x_j)$.
This leads to the objective of the optimization problem, being
\begin{equation}
    \sum_{i=1}^N \alpha_i - \frac{1}{2} \sum_{i, j = 1}^N \alpha_i \alpha_j y_i y_j k(x_i, x_j),
\end{equation}
and the decision function:
\begin{equation}
f(x) = \text{sgn} \left(\sum_{i = 0}^N y_i \alpha_i k(x, x_i) + b\right)
\end{equation}
for a specified kernel function $k$.

Although there are numerous kernels already deployed in SVMs in various applications~\cite{nalepa2019selecting,Tanveer2022}, the radial basis function (RBF) kernel is particularly widely used in SVMs~\cite{nalepa2020icpr}. The similarity measure for this kernel is given as:
\begin{equation}
    k(x_i, x_j) = e^{-\gamma || x_i - x_j ||^2}.
\end{equation}
The RBF kernel is known for its extremely high flexibility (its Vapnik–Chervonenkis \cite{vapnik1971chervonenkis, vapnik1999nature} dimension is infinite) and good generalization properties. Additionally, the RBF kernel is convenient to fine-tune, as it has only one parameter (the width of the kernel $\gamma$) which is commonly optimized together with the regularization parameter $C$. This is particularly important due to the high time and memory complexity of the SVM training, depending on the training set size. Hence, grid searching a large solution space may easily become infeasible to optimize the kernel hyperparameters. There are, however, fast approaches toward optimizing the training sets, kernel parameters and subsets of feature sets for SVMs which effectively exploit heuristic techniques to accelerate this process~\cite{nalepa2019selecting}.


\subsection{Quantum Kernel Methods}\label{sec:QKM}

The central motivation for utilizing quantum computational methods in SVM kernels is to take advantage of the exponentially large target space $\mathscr{H}$. This can lead to better separability of the data.
When considering the implementation of quantum kernel methods, a principal question that
quickly arises pertains to the way in which classical input data will be loaded into the quantum circuit.
In general, the objective will be to construct a unitary operator for each input datum $x$, such that applying it to the
initial quantum zero state leaves us with a specified representation of $x$.
Considering an example of a 1-qubit quantum circuit:
\begin{equation}
U_{\phi(x)} |0\rangle = |\phi(x)\rangle.
\end{equation}
This process is called \textit{quantum embedding}, while such transformation $U_{\phi(x)}$ induces \textit{quantum feature map} $\ket{\phi(x)}$.
In performing quantum embedding of a classical datum on $n$ qubits, we effectively map it into a $2^n$-dimensional Hilbert space:
\begin{equation}
    U_{\phi(x)} \ket{0}^{\otimes n} = \ket{\phi(x)} \in (\mathbb{C}^2)^{\otimes n} = \mathscr{H}.
\end{equation}

Some simpler, dimension-preserving examples of concrete feature maps may be recalled. One such method, which is referred to as the amplitude embedding, results in the quantum state with probability amplitudes corresponding to the components of the normalized input data vector. Let $x \in \mathbb{R}^n$, then its quantum-embedded form will become
\begin{equation}
    \ket{\psi} = \frac{1}{||x||} \sum_{i=1}^{2^n} x_i \ket{i},
\end{equation}
where $\ket{i}$ is the $i$-th Z-basis state.
Another noteworthy approach is basis embedding, which, in turn, considerably increases the dimension of the data, resulting in a state that is not in superposition. It builds on intuition brought by the analogy between classical binary sequences and corresponding z-basis states:
\begin{equation}
\begin{gathered}
    x \mapsto \ket{b_1 ... b_n}, \\
    x = [b_1 ... b_n]^T, \; b_i \in \{0, 1\}.
 \end{gathered}
\end{equation}

However, it is the ability to directly operate on complex high-dimensional data stored in qubits that makes quantum computing promising in the realm of data classification.
Therefore, commonly used feature maps aim to increase the dimensionality of input data while also exploring the possibilities provided by quantum entanglement and superposition.
Such methods of quantum embedding are introduced and discussed in Section \ref{sec:models}.

Considering a collection of quantum states obtained by means of applying a feature map to different classical input data, it is straightforward to reason about them in terms of kernel methods.
Kernel $K$ in regards to any two embedded classical data $x_1$, $x_2$ can be defined as the fidelity between the resulting quantum states:
\begin{equation}
\label{eq:quantum-kernel}
    K(x_i, x_j) = |\braket{\phi(x_i) | \phi(x_j)}|^2.
\end{equation}
Such kernel $K$ is known as a \textit{quantum kernel}, or a \textit{quantum embedding kernel} (QEK).

Taking into account that for any quantum state $\ket{\psi} \in \mathbb{C}^{2^n}$, only $\braket{0^n | \psi}$ could trivially be estimated
with the use of Z-basis measurement, a method for realizing the estimation presented in Equation~\ref{eq:quantum-kernel} needs to be
selected. A well-known approach would be to employ the \textit{swap test}, which can further be extended to allow fidelity estimation of two $n$-qubit states~\cite{prove2021extending}.
However, this comes with the requirement of having $3n$ qubits available: $n$ qubits for each of the quantum states being compared and $n$ ancilla qubits.

In a similar vein, a modification of the \textit{Hadamard test} can be made by extending the circuit with an $n$-qubit register and preceding the controlled application of $U(x_i)$
with the application of $U(x_j)$ to the new register, conditioned on the ancilla qubit being $\ket{0}$.
For such a circuit, the fidelity can be derived from measuring the final state of the ancilla qubit:
\begin{equation}
p(\ket{0}) = \frac{2 + 2 \operatorname{Re} \braket{U(x_j) | U(x_i)}}{4}.
\end{equation}
This approach reduces the number of required qubits to $2n + 1$ but, in turn, requires us to be able to construct the controlled version of $U(x)$, the unitary that embeds the classical datum $x$ into $n$ qubits.

\begin{figure}[!t]
\centering
\includegraphics[width=2.5in]{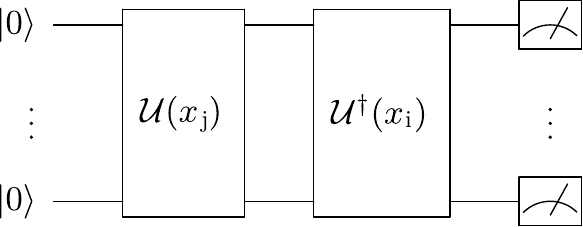}
\caption{Quantum circuit for estimating $\braket{U(x_i) | U(x_j)}$ (Equation~\ref{eq:Z_basis_QKE}), with $x_i, x_j$ embedded using an $n$-qubit $U$ operator, with the use of the circuit inversion method.}
\label{fig:QKE}
\end{figure}

Finally, if the state $\ket{\psi}$ is the result of applying $U(x_i) \ket{0^n}$, not unlike the ones in Havlicek's formulation of a hybrid SVM \cite{havlivcek2019supervised} employed in this work,
the fidelity between two states $\ket{U(x_i)}, \ket{U(x_j)}$ can be simply derived by concatenating to the existing circuit the hermitian conjugate of the transformation $U(x_j)$ and performing z-basis measurement on all qubits (Fig. \ref{fig:QKE}), yielding:

\begin{equation}\label{eq:Z_basis_QKE}
\braket{U(x_j) | U(x_i)} = P(\ket{0^n}) = \braket{0^n | U^{\dag}(x_j) U(x_i) | 0^n}.
\end{equation}

\begin{figure*}[!t]
\centering
\subfloat(a){\includegraphics[height=1. in]{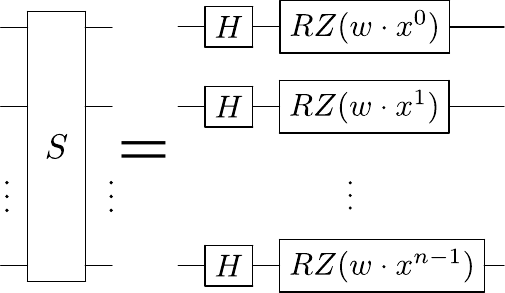}%
\label{fig:S}}
\subfloat(b){\includegraphics[height=1. in]{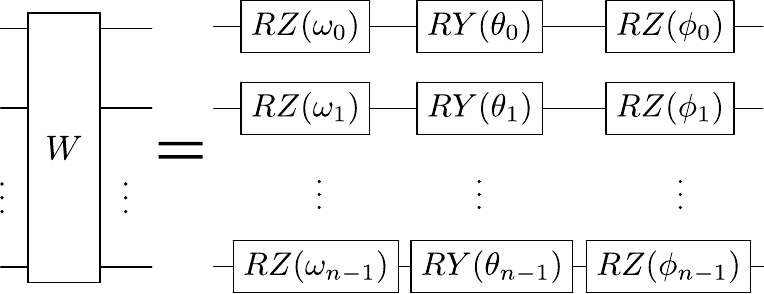}%
\label{fig:W}}
\subfloat(c){\includegraphics[height=1. in]{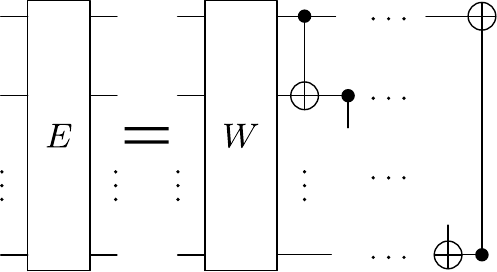}%
\label{fig:E}}
\caption{Layers with which we construct circuit architectures. a) Data encoding layer $S$. b) Variational layer $W$ c) Entangling variational layer $E$. For discussion of the layers refer to the Sec. \ref{sec:models}}
\label{fig:layers}
\end{figure*}

\subsection{Kernel Target Alignment}\label{sec:KTA}
Using blindly a huge size of the target Hilbert space $\mathscr{H}$ in order to rigidly embed the classical data can backfire on the resulting classification performance. 
Firstly, one can easily overfit the model leading to the poor generalization performance.
Secondly, in the high dimensional space, almost all vectors are orthogonal to each other, causing the vanishing of fidelity based kernels (Equation~\ref{eq:quantum-kernel}) and possible untrainability of the learning models \cite{thanasilp2022exponential}.
Therefore, one should look for a trade off between the size of the quantum embedded feature space and the above obstructions. It has been proposed \cite{hubregtsen2022training ,thanasilp2022exponential} to introduce additional, variational hyperparameters to the quantum feature maps to calibrate them for the specific learning task. Those hyperparameters can be chosen by maximizing a function called Kernel Target Alignment, which we introduce below.

For a given set of data $\{x_1, \dots, x_N\}$ a kernel function can be represented through a Gram matrix:
\begin{equation}
    K(x_i, x_j) = K_{ij}.
\end{equation}
Each entry in the above matrix indicates the mutual similarity for the data points $x_i$ and $x_j$. Consider a kernel function:
\begin{equation*}
    \bar{K}(x_i, x_j)= \begin{cases}
    +1 &\text{if $x_i$ and $x_j$ are in the same class}\\
    -1 &\text{if $x_i$ and $x_j$ are in different classes}.
    \end{cases}
\end{equation*}
It shows a clear distinction between classes of data points. 
If one could construct a feature map that gives rise to the above kernel function, then one would obtain the perfect SVM performance. Therefore, $\bar{K}$ is called the \textit{ideal} kernel. As SVMs are supervised learning models, for a given training data, one can use data point labels to construct the \textit{ideal} kernel matrix,
\begin{equation}\label{eq:ideal}
\bar{K}_{ij} = y_i y_j,
\end{equation}
where $y_i, y_j \in \{+1,-1\}$ are the labels of the data points $x_i, x_j$. In general, in almost every situation, one will not be able to find the exact feature map, which gives rise to the ideal kernel. 
Therefore, parametrized families of feature maps are used to optimize the resulting kernel matrix in such a way that it resembles the ideal kernel as closely as possible.

To compare two kernel matrices, one can use the matrix alignment given as:
\begin{equation}\label{eq:matrix_alignment}
    \mathcal{A}(K_1, K_2) = \frac{\langle K_1, K_2 \rangle_F}{\sqrt{\langle K_1, K_1 \rangle_F \langle K_2, K_2 \rangle_F}},
\end{equation}
where $\langle K_1, K_2\rangle_F = Tr\{ K_1^T K_2 \}$ is a Frobenius inner product.
One can utilize the matrix alignment $\mathcal{A}$ to create a smooth function of kernel function parameters, which measures the similarity between the specific and ideal kernel matrices. It is called Kernel Target alignment:
\begin{equation}\label{eq:target_alignment}
\mathcal{T}(K) = \mathcal{A}(\bar{K}, K) = \frac{\sum_{ij} y_i y_j K_{ij}}{ \sqrt{ \left( \sum_{ij} K_{ij}^2 \right) \left( \sum_{ij}y_i^2 y_j^2 \right) } }.
\end{equation}
As expected, the Kernel-Target alignment correlates with the performance of the classifier \cite{cristianini2001kernel, wangKernelAlignment}, and it is commonly used in the model selection process. 
Denoting kernel families obtained from the parameterized feature maps as $K(\theta)$, where $\theta$ is a hyperparameter (or a set of hyperparameters), we can express the kernel optimization task as:
\begin{equation}\label{eq:TAoptimization}
    \max_{\theta}\ \mathcal{T}(K(\theta)).
\end{equation}

\section{Methods}\label{sec:methods}

This section presents the introduced hybrid models which are exploited for multispectral data analysis (Section~\ref{sec:models}). Additionally, we discuss our approach for reducing the SVM training sets in Section~\ref{sec:superpixel_reduction}---this step is pivotal to enable us to train SVMs from massively large Earth observation data.

\subsection{Hybrid Models}\label{sec:models}

In this work, we introduce the circuit architectures for kernel estimation in the cloud classification task. Those circuits are designed with three types of gate layers (Fig.~\ref{fig:layers}):
\begin{itemize}
    \item \textbf{Data encoding layer $S$}---The rotations by the $w$-rescaled value of the specific feature performed on the corresponding qubits. We keep the same scaling factor for each feature $w = \pi$. The initial layer of Hadamard gates is introduced in order utilize the superposition of states by abandoning the computational Z-basis. With such a map, each feature is encoded into different quantum register, therefore the number of features ultimately equals the number of qubit registers $n=m$.
    \item \textbf{Variational layer $W$}---the parameterized arbitrary rotations of each qubit. Each $W$ layer introduces $3 \cdot n$ variational hyperparameters. The layer is implemented by broadcasting {\fontfamily{cmtt}\selectfont Pennylane}'s {\fontfamily{cmtt}\selectfont Rot} operation. Both $S$ and $W$ layers consist of one-qubit gates, hence they do not introduce entanglement to the system. There is a perfect separation of the qubit registers in the circuit.
    \item \textbf{Entangling variational layer $E$}---the $W$ layer with strong entangling of qubits done by CNOT two-qubit gates. The layer is implemented by the {\fontfamily{cmtt}\selectfont Pennylane}'s {\fontfamily{cmtt}\selectfont StronglyEntanglingLayers} operation.
\end{itemize}

Our circuit architectures are recognized by their layer composition. 
By using the $WS_n$ symbols, we mean that the data embedding map first 
transforms the initial state $| 0 \rangle^{\otimes n}$ by the arbitrary 
rotations layer $W$, then the data is encoded with the $S$ layer. To 
estimate a quantum kernel entry, the conjugate embedding map with respect 
to a different data point is concatenated to the $WS_n$ circuit, 
as explained in Section~\ref{sec:QKM}. Other architecture symbols follow 
the same rule.

In this study, we investigate the $S$, $WS$, $ES$, $WSWS$ circuit architectures. 
This choice enables us to analyze the significance of hyperparameter tuning, superposition, entanglement, and expressivity in quantum feature maps, while being able to simulate them with classical computers on a real-life dataset. It is worth emphasizing that for the $S$ and $W$ layers-based maps the quantum kernel complexity is expected to be low. Such circuits are, therefore, easy to simulate on classical computers, and the application of quantum computers does not provide an advantage here. However, precisely thanks to this property, we were able to perform studies for real-world large-scale datasets, which would be much more difficult to do in the case of more complex quantum kernels. This especially concerns the ZZ map discussed in~\cite{havlivcek2019supervised,goldberg2017complexity}, for which the computations are $\# P-$hard for classical computers. The case of the ZZ map was beyond the reach of our computational abilities for the complex dataset under investigation. However, we managed to analyze 
intermediate complexity kernels involving the entangling $E$ layer. Hence, in this article, we focus on applying the quantum-kernel methods to huge amounts of real Earth observation data captured in orbit. However, this was achieved by the cost of reducing the kernels' complexity. 

\begin{figure}[!t]
\centering
\includegraphics[width=2.5in]{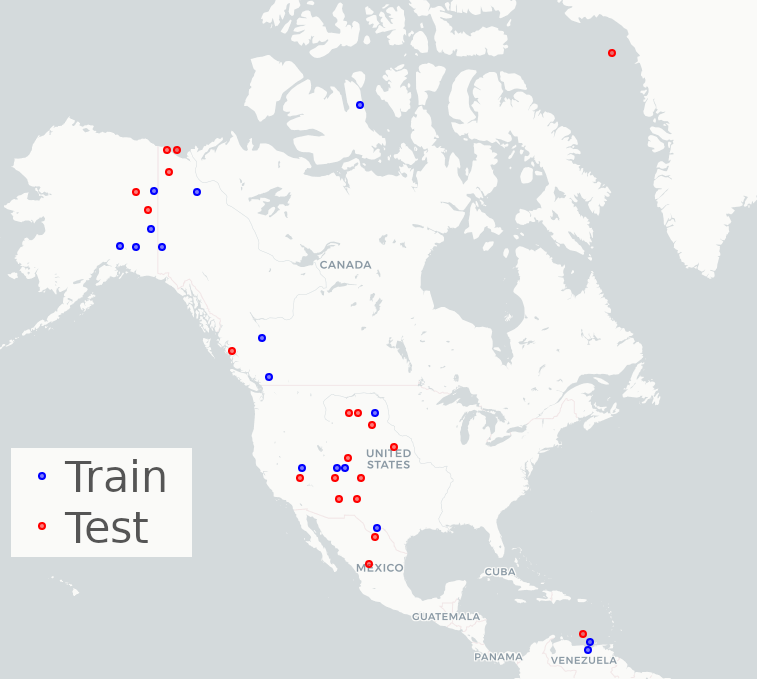}
\caption{The location of each 38-Clouds dataset training and test scene.}
\label{fig:map}
\end{figure}

\subsection{Training Data Reduction}\label{sec:superpixel_reduction}

\begin{figure}[!t]
\centering
\includegraphics[width=2.5in]{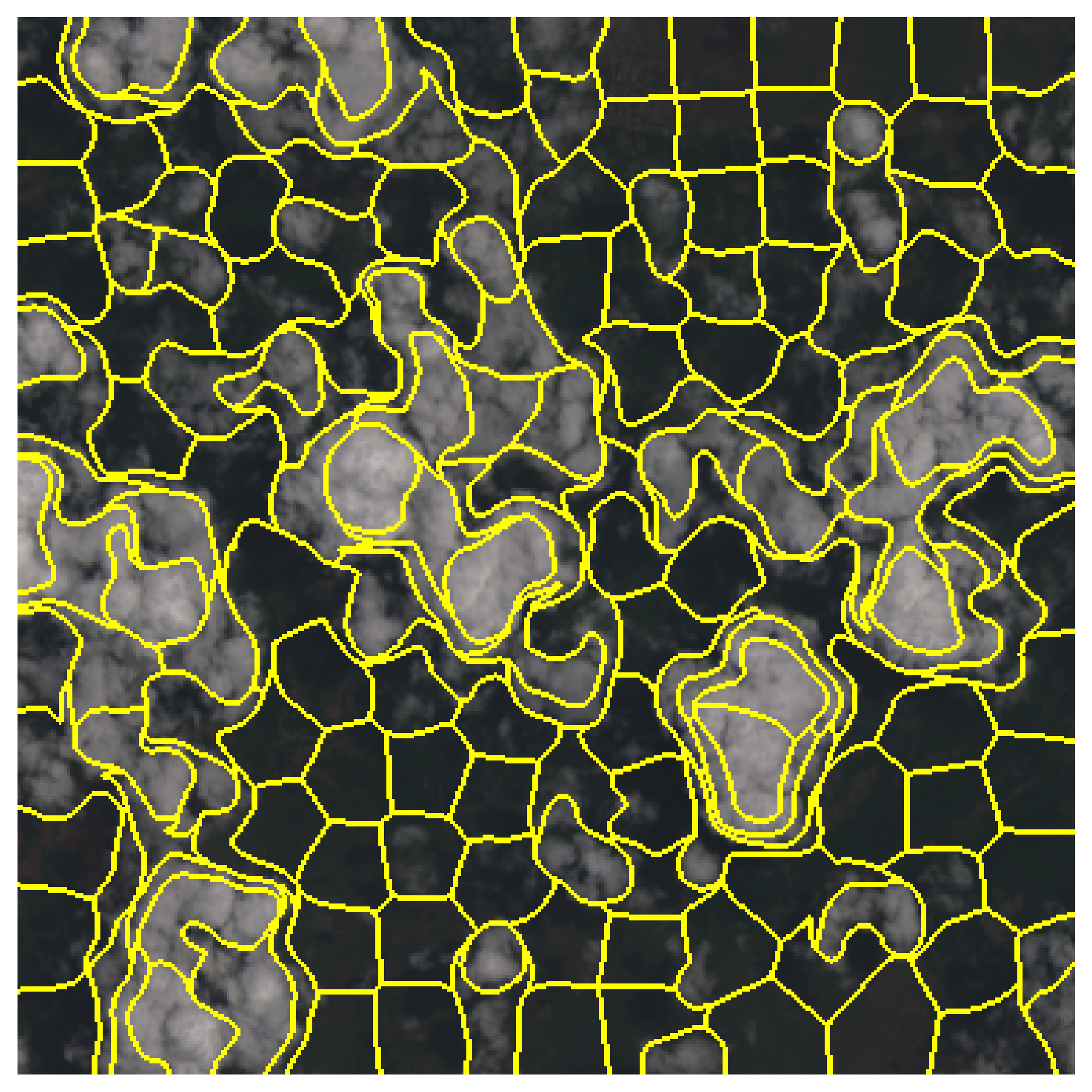}
\caption{SLIC superpixel segmentation applied to the 38-Cloud training patch. Yellow lines indicate the borders of the segment.}
\label{fig:SLIC}
\end{figure}

SVMs suffer from their high time and memory training complexity, which depend on the size of the training set. 
The 38-Clouds training data consists of approximately $1.24$ billion of pixels (the dataset is presented in more detail in Section~\ref{sec:dataset}), the use of all of them is implausible as this size significantly exceeds the computational capabilities of modern computers in the context of the SVM training. Since only a subset of all training vectors is annotated as support vectors during the process of SVM training, we can effectively exploit only a subset of the most important examples or create the \textit{prototype vectors} which are a good representation of similar examples \cite{nalepa2019selecting} (e.g., combining the information captured by several neighboring vectors in the feature space). In this work, we follow the later approach by utilizing the superpixel segmentation techiques~\cite{ren2003learning}. 
Here, we create coherent pixel groupings by considering similarity measures defined using perceptual features---we build upon the famous Simple Linear Iterative Clustering (SLIC) \cite{achanta2010slic} which performs the segmentation based on color and proximity distance (see an example result of SLIC obtained for the 38-Cloud data sample rendered in Fig.~\ref{fig:SLIC}). For each multispectral training patch, we do the following steps:
\begin{enumerate}
    \item Perform SLIC ($N_{segments} = 200$, smoothing kernel $\sigma = 5$; the hyperparameters of SLIC were fine-tuned experimentally, in order to compromise between the reduction rate and the spatial representativeness of the resulting training examples) segmentation.
    \item Remove margin pixels from each segment.
    \item Create a prototype training vector example from each segment by computing the following statistical measures for each spectral band: mean, median, interquartile range, min, max, standard deviation.
    \item Label the created superpixel with the majority label of the pixels contained within the corresponding superpixel.
\end{enumerate}
After data reduction, we obtained approximately $0.93$ million of training vectors (resulting in the massive reduction rate of more than $1300\times$) consisting of 24 features and a ground-truth label. Additionally, almost $92\%$ of superpixels have their class label decided by at least 80-20 vote ratio.

\section{Experimental Validation}\label{sec:validation}

In this section, we discuss our experimental study focused on understanding the abilities of hybrid SVMs in the context of cloud detection in satellite multispectral data. The exploited dataset is discussed in detail in Section~\ref{sec:dataset}, whereas the experimental methodology is highlighted in Section~\ref{sec:methodology}. The results are presented and discussed in Section~\ref{sec:results}.

\subsection{Dataset}\label{sec:dataset}

We utilize satellite multispectral image data contained in the 38-Cloud dataset~\cite{38-cloud-1,38-cloud-2}. It consists of 18 training and 20 test scene images captured by the Landsat-8 satellite (30\,m ground sampling distance) over the continent of America (Fig. \ref{fig:map}). Scenes cover a wide range of climate zones and terrain types, including deserts, forests, meadows, mountains, agriculture, urban areas, coastlines, snow, and ice. With each scene, we are provided the ground truth for cloud binary classification. There is no gradation in the cloud labels; hence, this class includes both thick cumulus, partly transparent cirrus clouds as well as thin haze. For convenience, scene images are cropped into $8400$ (training) and $9201$ (test) $384 \times 384$ pixel patches by the authors of the dataset. Each pixel has five values associated with it: intensity values in four spectral bands (blue: $450-515$\,nm, green: $520-600$\,nm, red: $630-680$\,nm, NIR: $845-885$\,nm) and a ground-truth label (cloud or background). 
It is worth noting that the scene images are not rotated to fit the standard rectangular image format, therefore, they include a significant amount of margin pixels, represented by $[0,0,0,0]$ vectors with the non-cloud (background) class label assigned.

\subsection{Experiment Methodology}\label{sec:methodology}

We investigate the performance of classical and hybrid learning models based on SVMs. The classical SVMs are trained and tested for the RBF and linear kernel. Hybrid, classical-quantum models consist of two parts, the quantum kernel estimation (QKE) and the classical SVM routine. The evaluated models are trained on partly random balanced samples from the reduced training set obtained by the procedure described in Section \ref{sec:superpixel_reduction}. The training data sample of size $N$ is obtained by randomly selecting $\frac{N}{2}$ superpixels which have a \textit{cloud} label and $(\frac{N}{2}-1)$ superpixels corresponding to a \textit{non-cloud} label. Then, the last ($N^{th}$) \textit{non-cloud} superpixel is added---it contains zeroed features and represents the ``margin'' superpixel. Due to the substantial number of margin pixels in each scene in the 38-Cloud dataset, not including a ``margin" training example could result in the significant drop in the model's performance. We evaluated all models on a fixed set of training samples. For each training set size $N \in \{10, 20, 40, 80, 160, 320, 640, 1280\}$, we randomly sample $20$ sets (which remain unchanged across all investigated SVM models). Hence, each model is trained $160$ times. The investigated hybrid approaches were implemented in the {\fontfamily{cmtt}\selectfont Pennylane} python package, and the experiments were run with the {\fontfamily{cmtt}\selectfont default.qubit} simulator on classical computers.

The dimension of the quantum Hilbert space, to which we encode the data, grows exponentially with the number of qubits $n$ that we use in quantum feature maps. All maps that we use encompass exactly the same amount of qubits as the number of features $m$, hence $n=m$. The dimension of Hilbert space needed to encode superpixels in the training data set is $2^{24} (= 16,777,216)$. Performing large-scale simulations of such big space on modern classical computers is unfeasible. Therefore, in the case of the hybrid models, we perform feature extraction by Principal Component Analysis (PCA). In the vast majority of cases, extracting two (four) principal components explains $95\%$ ($99\%$) of variance in the training data. 

With each training set, we draw an additional balanced validation set of size $\max(N/2, 300)$.
For the classical SVM with linear kernel and the hybrid $S_4$ model, we tuned only the hyperparameter $C$. It was done by testing different classifiers on a validation sample in the hyperparameter range $C \in [0.01,147.01]$ with the step size of $3$. For the classical SVM with the RBF kernel, the procedure is the same, but two hyperparameters ($C, \gamma$) were tuned (the ranges of hyperparameters stay the same).
In the case of hybrid $WS_4, ES_4$ ($WSWS_4$) models, we first fine-tuned their $12$ ($24$) variational layer parameters by optimizing the Kernel Target alignment $\mathcal{T}(K)$ (Equation \ref{eq:target_alignment}) on the training sample. For this task, we used the Adam optimizer~\cite{kingma2014adam}. Once the variational layers have been fine-tuned, we performed $C$ hyperparameter tuning in the same manner as for other classifiers.

The learning models are evaluated on all $20$ test scenes in the 38-Cloud dataset. It means that each model is evaluated 3,200 times ($20$ training samples $\times$ $8$ training sample sizes $\times$ $20$ test scenes). To quantify the performance of the investigated classification models, we exploited accuracy: $\Accuracy = {({\rm TP} + {\rm TN})}/{({\rm TP} + {\rm TN} + {\rm FN} + {\rm FP})}$, Jaccard index: $\Jaccard = {\rm TP}/({\rm TP}+{\rm FN}+{\rm FP})$, precision: $\Precision = {{\rm TP}}/{({\rm TP} + {\rm FP})}$, recall: $\Recall = {{\rm TP}}/{({\rm TP} + {\rm FN})}$, and specificity: $\Specificity={\rm TN}/({\rm TN}+{\rm FP})$, where $\rm TP$, $\rm TN$, $\rm FP$ and $\rm FP$ denote true positives, true negatives, false positives and false negatives, respectively. All results are reported for the test sets that were unseen during training (unless stated otherwise).

\begin{table*}[ht!]
\centering
\renewcommand{\tabcolsep}{7mm}
\caption{The results for $RBF_4$ classical model, as well as for the $WS_4$ and $WSWS_4$ hybrid models. The best results are boldfaced for each model---we report the average (standard deviation) of the corresponding metric obtained across 20 independent executions for each size of the refined training set.}
\begin{tabular}{rrrrrr}

\hline
{N}    & \multicolumn{1}{c}{\Accuracy} & \multicolumn{1}{c}{$\Jaccard$} & \multicolumn{1}{c}{\Precision} & \multicolumn{1}{c}{\Recall} & \multicolumn{1}{c}{\Specificity} \\
\hline
\multicolumn{6}{c}{The $RBF_4$ model}\\
\hline
\textbf{10}   & 0.820 (0.138)      & 0.515 (0.123)     & 0.631 (0.132)       & 0.744 (0.177)    & 0.841 (0.201)         \\
\textbf{20}   & 0.849 (0.084)      & 0.525 (0.105)     & 0.652 (0.089)       & 0.721 (0.163)    & 0.889 (0.086)         \\
\textbf{40}   & 0.872 (0.067)      & 0.550 (0.084)     & 0.680 (0.088)       & 0.712 (0.130)    & 0.913 (0.076)         \\
\textbf{80}   & 0.884 (0.054)      & 0.576 (0.075)     & 0.685 (0.080)       & 0.754 (0.075)    & 0.919 (0.061)         \\
\textbf{160}  & 0.898 (0.036)      & 0.581 (0.068)     & 0.701 (0.063)       & 0.743 (0.080)    & 0.939 (0.036)         \\
\textbf{320}  & 0.907 (0.026)      & 0.606 (0.058)     & 0.729 (0.059)       & 0.757 (0.060)    & 0.948 (0.029)         \\
\textbf{640}  & 0.911 (0.020)      & 0.609 (0.047)     & 0.732 (0.048)       & \textbf{0.763 (0.040)}    & 0.954 (0.019)         \\
\textbf{1280} & \textbf{0.919 (0.010)}      & \textbf{0.628 (0.035)}     & \textbf{0.757 (0.036)}       & \textbf{0.763 (0.029)}    & \textbf{0.963 (0.009)}         \\
\hline
\multicolumn{6}{c}{The $WS_4$ hybrid model}\\
\hline
\textbf{10}   & 0.769 (0.156)      & 0.446 (0.157)     & 0.599 (0.131)       & 0.684 (0.245)    & 0.810 (0.213)          \\
\textbf{20}   & 0.821 (0.128)      & 0.503 (0.119)     & 0.621 (0.125)       & 0.717 (0.178)    & 0.847 (0.191)         \\
\textbf{40}   & 0.838 (0.128)      & 0.530 (0.112)     & 0.642 (0.130)       & 0.737 (0.142)    & 0.858 (0.196)         \\
\textbf{80}   & 0.870 (0.066)      & 0.554 (0.091)     & 0.675 (0.094)       & 0.737 (0.119)    & 0.905 (0.070)          \\
\textbf{160}  & 0.881 (0.057)      & 0.553 (0.096)     & 0.708 (0.097)       & 0.703 (0.135)    & 0.929 (0.055)         \\
\textbf{320}  & 0.893 (0.044)      & 0.568 (0.092)     & 0.709 (0.088)       & 0.731 (0.122)    & 0.935 (0.036)         \\
\textbf{640}  & 0.895 (0.053)      & 0.573 (0.100)     & 0.717 (0.087)       & 0.731 (0.129)    & 0.940 (0.040)        \\
\textbf{1280} & \textbf{0.911 (0.031)}      & \textbf{0.602 (0.064)}     & \textbf{0.723 (0.061)}       & \textbf{0.757 (0.076) }   & \textbf{0.949 (0.026)}        \\
\hline
\multicolumn{6}{c}{The $WSWS_4$ hybrid model}\\
\hline
\textbf{10}   & 0.770 (0.156)      & 0.445 (0.159)     & 0.600 (0.132)       & 0.683 (0.247)    & 0.811 (0.213)         \\
\textbf{20}   & 0.822 (0.127)      & 0.505 (0.117)     & 0.622 (0.124)       & 0.718 (0.175)    & 0.847 (0.191)         \\
\textbf{40}   & 0.856 (0.082)      & 0.541 (0.095)     & 0.654 (0.089)       & 0.737 (0.144)    & 0.888 (0.088)         \\
\textbf{80}   & 0.870 (0.067)      & 0.554 (0.092)     & 0.674 (0.093)       & 0.737 (0.120)    & 0.904 (0.071)         \\
\textbf{160}  & 0.881 (0.059)      & 0.552 (0.096)     & 0.710 (0.096)       & 0.703 (0.140)    & 0.930 (0.058)         \\
\textbf{320}  & 0.894 (0.044)      & 0.575 (0.082)     & 0.698 (0.081)       & 0.744 (0.104)    & 0.931 (0.038)        \\
\textbf{640}  & 0.899 (0.048)      & 0.582 (0.089)     & 0.714 (0.081)       & 0.745 (0.114)    & 0.940 (0.038)         \\
\textbf{1280} & \textbf{0.910 (0.031)}      & \textbf{0.602 (0.064)}     & \textbf{0.723 (0.062)}       & \textbf{0.758 (0.076)}    & \textbf{0.949 (0.026)}        \\
\hline
\end{tabular}
\label{tab:RBF4}
\end{table*}

\subsection{Results}\label{sec:results}

\begin{figure}[ht!]
\centering
\includegraphics[width=3in]{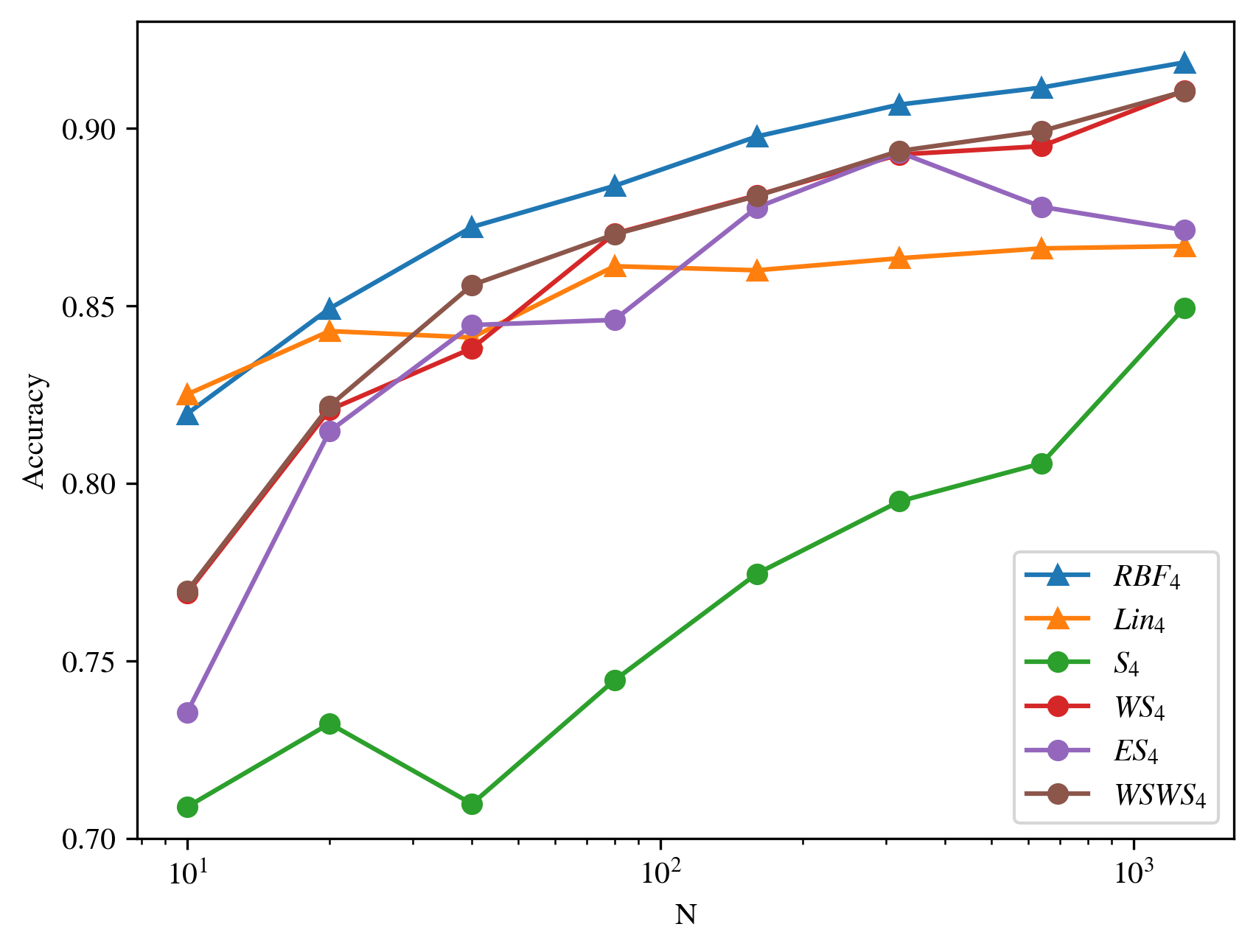}
\caption{The overall test accuracy (\Accuracy) of learning models for different training sample sizes $N$. The classical SVM models are shown with triangular markers, while hybrid models are shown with circular markers.}
\label{fig:Acc_N}
\end{figure}

\begin{figure}[ht!]
\centering
\includegraphics[width=0.5\textwidth]{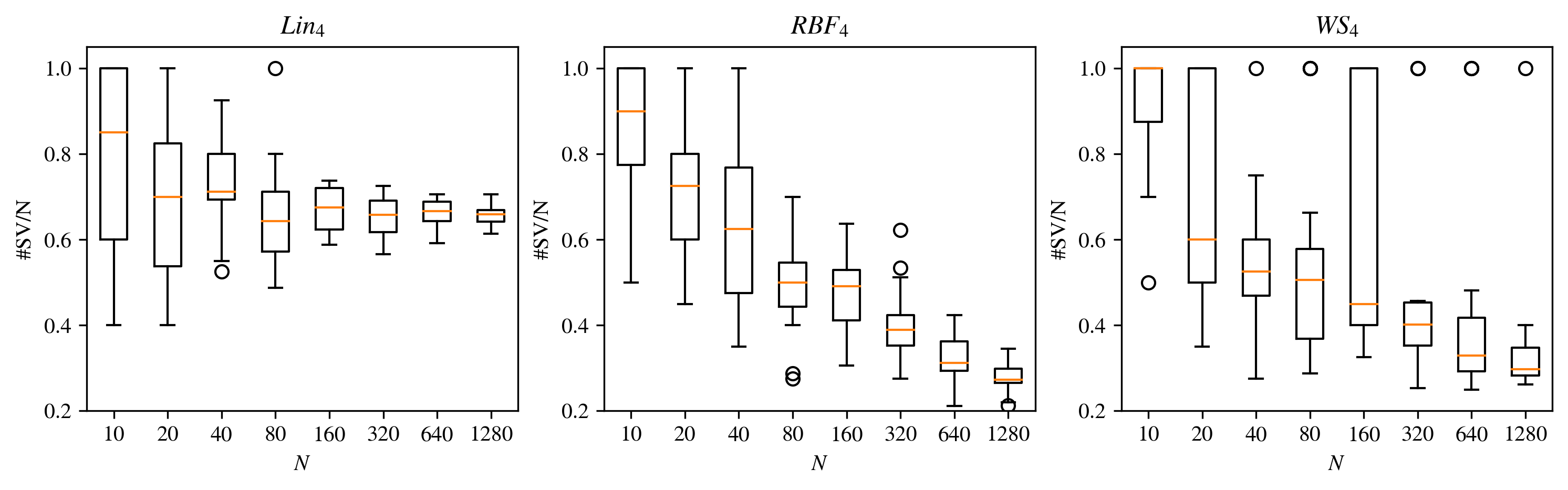}
\caption{The ratio of the number of support vectors (\#SV) and the size of the training set (N) obtained for the $Lin_4$, $RBF_4$, and $WS_4$ SVM models over all 20 independent executions for each $N$. }
\label{fig:Svs}
\end{figure}

The objectives of our experimental study is two-fold: (\textit{i})~to understand the impact of an increasing training set size on the generalization capabilities of both classical and hybrid SVMs, and (\textit{ii})~to investigate the performance of the proposed quantum classifiers in a real-world Earth observation task of cloud detection from multispectral imagery.  In Fig.~\ref{fig:Acc_N}, we render accuracy (averaged across all independent executions for each training set size) for all models. We can observe that increasing the size of the reduced training sets leads to the consistent increase in the classification performance of all SVM models. It is of note that the rate of the performance increase started saturating for the RBF SVM model ($RBF_4$), whereas the quantum-kernel classifiers ($WS_4$ and $WSWS_4$) manifest more rapid improvements for larger N's. This phenomenon can be further investigated in Table~\ref{tab:RBF4}, we gather all quantitative metrics obtained using the best classical SVM with the RBF kernel ($RBF_4$), together with our quantum SVMs. Finally, in Fig.~\ref{fig:Svs}, we present the ratio of the number of support vectors elaborated during the training process of the underlying model ($Lin_4$, $RBF_4$, and $WS_4$). Since the inference time of SVMs depends linearly on the number of support vectors, their number should be minimized to ensure fast operation of the classifier. Although there are indeed outlying executions resulting in large numbers of SVs for the $WS_4$ SVMs, the overall trend in the number of SVs remains consistent for all $N$'s (see the median number of SVs rendered as orange lines in Fig~\ref{fig:Svs}). 

\begin{figure*}[!t]
\centering
\includegraphics[width=0.92\textwidth]{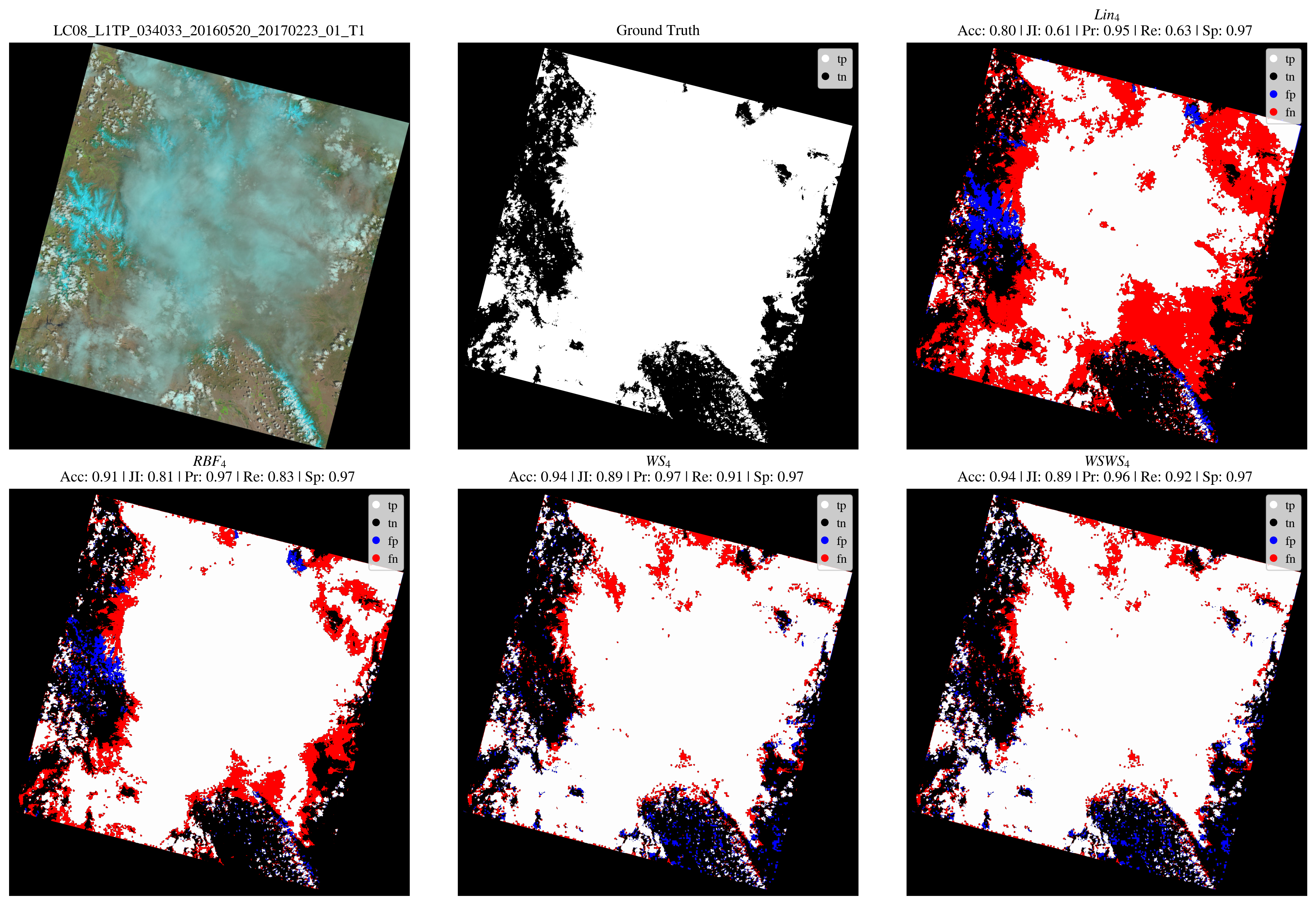}
\caption{The visualization of predictions of different models ($Lin_4$, $RBF_4$, $WS_4$, $WSWS_4$), alongside the quantitative metrics. All models were trained on one of the training samples of size $N=1280$ (the same for all classifiers), together with the natural false color scene image (here: LC08\_L1TP\_034033\_20160520\_20170223\_01\_T1), and the ground truth corresponding to this scene.}
\label{fig:LC08_L1TP_034033_20160520_20170223_01_T1}
\end{figure*}

\begin{figure*}[!t]
\centering
\includegraphics[width=0.92\textwidth]{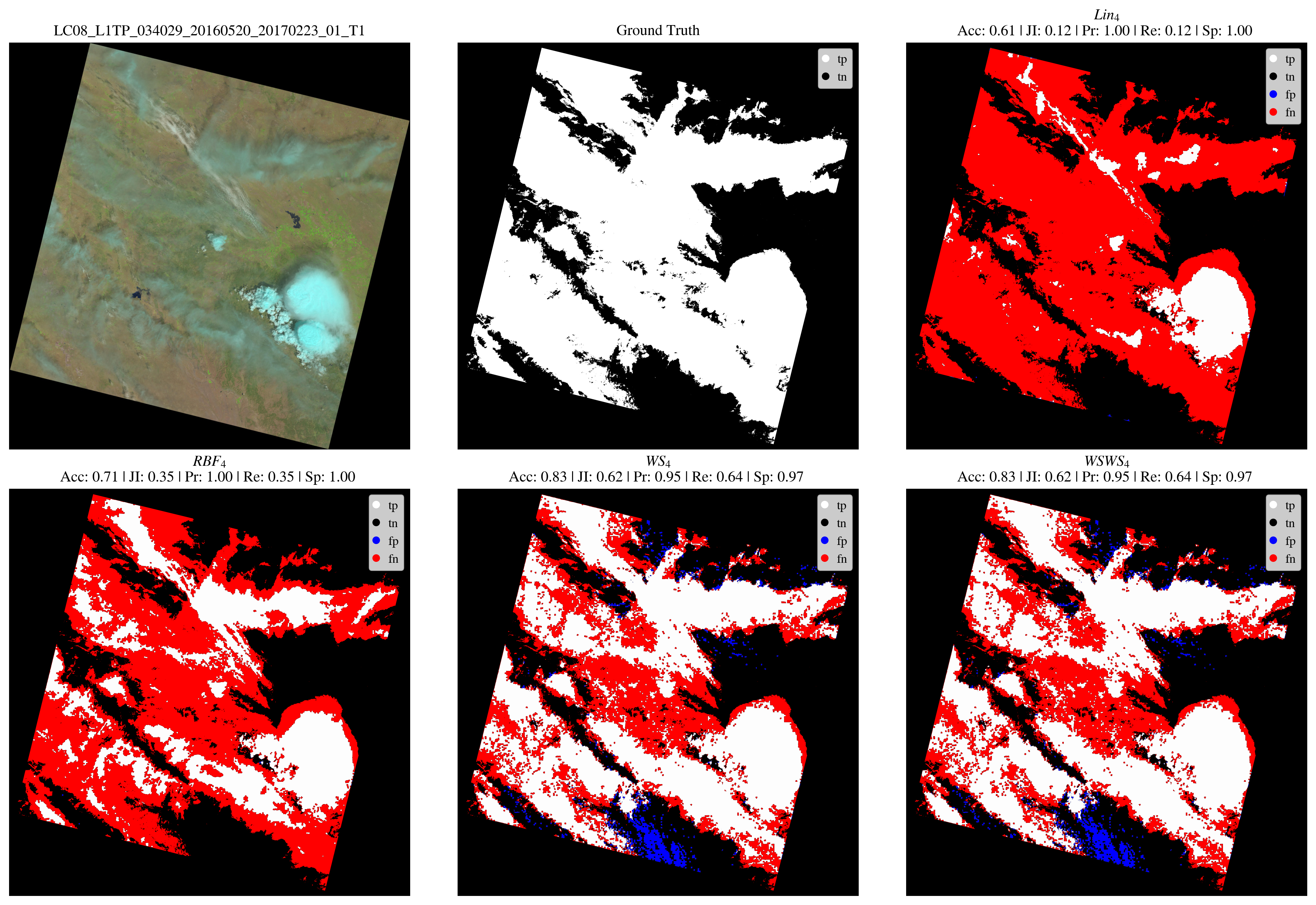}
\caption{The visualization of predictions of different models ($Lin_4$, $RBF_4$, $WS_4$, $WSWS_4$), alongside the quantitative metrics. All models were trained on one of the training samples of size $N=1280$ (the same for all classifiers), together with the natural false color scene image (here: LC08\_L1TP\_034029\_20160520\_20170223\_01\_T1), and the ground truth corresponding to this scene.}
\label{fig:LC08_L1TP_034029_20160520_20170223_01_T1}
\end{figure*}

\begin{figure*}[!t]
\centering
\includegraphics[width=0.92\textwidth]{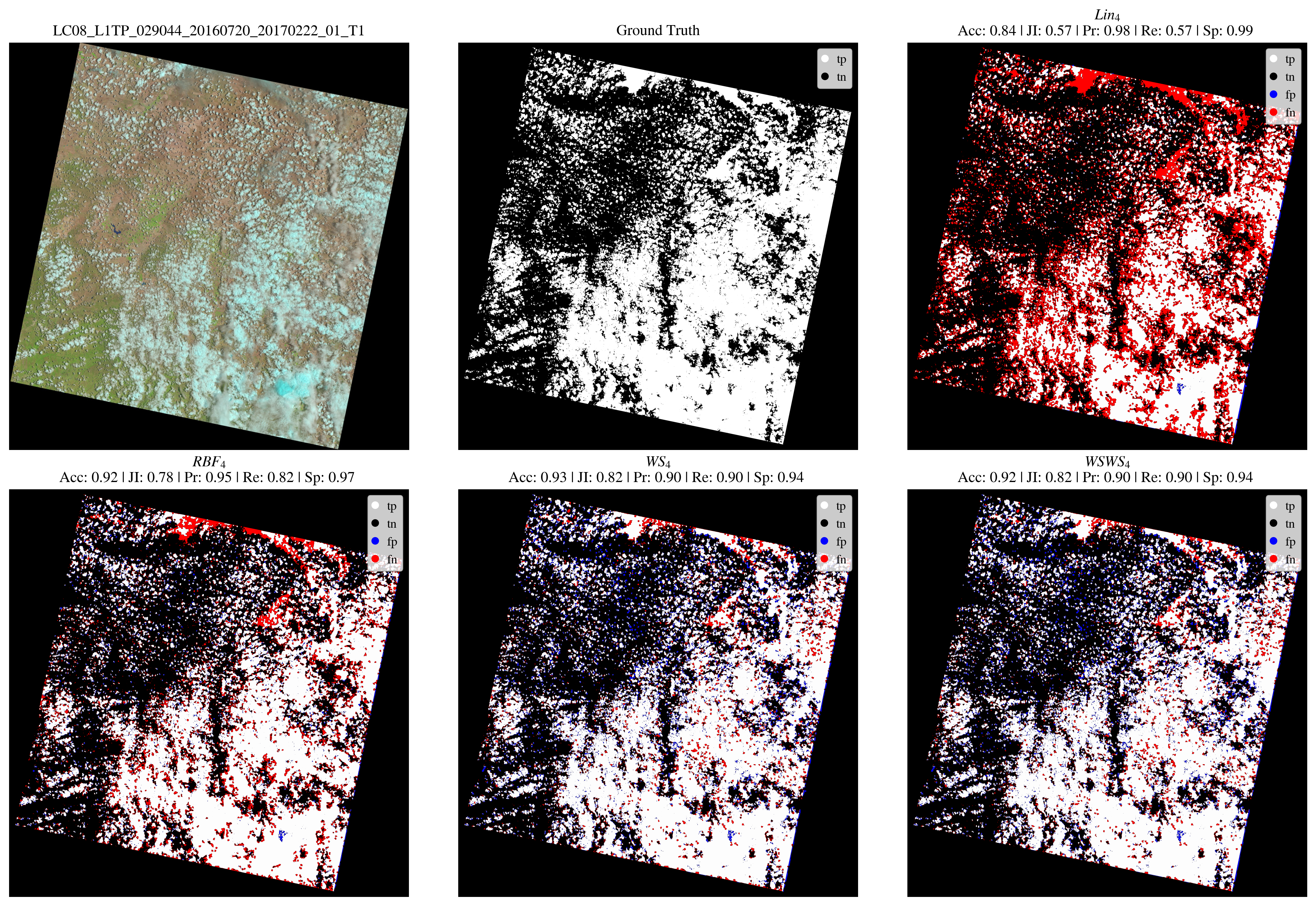}
\caption{The visualization of predictions of different models ($Lin_4$, $RBF_4$, $WS_4$, $WSWS_4$), alongside the quantitative metrics. All models were trained on one of the training samples of size $N=1280$ (the same for all classifiers), together with the natural false color scene image (here: LC08\_L1TP\_029044\_20160720\_20170222\_01\_T1), and the ground truth corresponding to this scene.}
\label{fig:LC08_L1TP_029044_20160720_20170222_01_T1}
\end{figure*}

To verify if the differences across the investigated models are statistically important, we executed the Friedman's tests with post-hoc Dunn's over all metrics, averaged across all independent executions for the sampled refined training sets (Table~\ref{tab:dunns}). We can appreciate that the $WS_4$ and $WSWS_4$ models, with the former being significantly less parameterized than the latter one, lead to statistically same cloud detection performance. Additionally, once the dataset is increased and reaches the size of $N=1280$, the quantum-kernel SVMs deliver statistically-same quality measures as $RBF_4$. In Figs.~\ref{fig:LC08_L1TP_034033_20160520_20170223_01_T1}--\ref{fig:LC08_L1TP_029044_20160720_20170222_01_T1}, we present three example 38-Cloud test scenes of varying segmentation difficulty (see different cloud characteristics). The qualitative analysis shows that the quantum-kernel SVMs can indeed outperform or work on par with well-established SVMs with the RBF kernel, and both of them significantly outperform linear-kernel classifiers in this task. 

\begin{table*}[ht!]
\caption{
Adjusted P-values for Friedman's with post-hoc Dunn's multiple comparisons tests for the evaluation metrics elaborated using the $Lin_4$, $RBF_4$, $WS_4$, $WSWS_4$ models. The background of the statistically significant ($p<0.05$) results is grayed.}
\label{tab:dunns}
\centering
\begin{tabular}{rcccccc}
\hline
{N}    & \textbf{$Lin_4{\rm~vs.~}RBF_4$} & \textbf{$Lin_4{\rm~vs.~}WS_4$}                       & \textbf{$Lin_4{\rm~vs.~}WSWS_4$}          & \textbf{$RBF_4{\rm~vs.~}WS_4$}                       & \textbf{$RBF_4{\rm~vs.~}WSWS_4$}          & \textbf{$WS_4{\rm~vs.~}WSWS_4$}           \\ \hline
\multicolumn{7}{c}{\Accuracy}\\
\hline
\textbf{10}   & \textgreater{}0.9999 & \cellcolor[HTML]{E7E6E6}0.0087               & \cellcolor[HTML]{E7E6E6}0.0132               & \cellcolor[HTML]{E7E6E6}0.0023               & \cellcolor[HTML]{E7E6E6}0.0036               & \textgreater{}0.9999 \\
\textbf{20}   & 0.8499               & \cellcolor[HTML]{E7E6E6}0.0087               & 0.1649               & \cellcolor[HTML]{E7E6E6}\textless{}0.0001    & \cellcolor[HTML]{E7E6E6}0.0014               & \textgreater{}0.9999 \\
\textbf{40}   & \cellcolor[HTML]{E7E6E6}0.0423               & 0.0858               & \textgreater{}0.9999 & \cellcolor[HTML]{E7E6E6}\textless{}0.0001    & 0.3003               & \cellcolor[HTML]{E7E6E6}0.0087               \\
\textbf{80}   & \cellcolor[HTML]{E7E6E6}0.0057               & 0.8499               & \textgreater{}0.9999 & 0.3972               & \cellcolor[HTML]{E7E6E6}0.0423               & \textgreater{}0.9999 \\
\textbf{160}  & \cellcolor[HTML]{E7E6E6}\textless{}0.0001    & 0.0607               & 0.1649               & 0.0858               & \cellcolor[HTML]{E7E6E6}0.0291               & \textgreater{}0.9999 \\
\textbf{320}  & \cellcolor[HTML]{E7E6E6}0.0023               & 0.5185               & 0.0607               & 0.3972               & \textgreater{}0.9999 & \textgreater{}0.9999 \\
\textbf{640}  & \cellcolor[HTML]{E7E6E6}0.0291               & \textgreater{}0.9999 & \textgreater{}0.9999 & \cellcolor[HTML]{E7E6E6}0.0197               & 0.6681               & \textgreater{}0.9999 \\
\textbf{1280} & 0.0858               & 0.1649               & 0.5185               & \textgreater{}0.9999 & \textgreater{}0.9999 & \textgreater{}0.9999\\ \hline
\multicolumn{7}{c}{$\Jaccard$}\\
\hline
\textbf{10}   & \textgreater{}0.9999 & \cellcolor[HTML]{E7E6E6}0.0014               & 0.1649               & \cellcolor[HTML]{E7E6E6}0.0423               & \cellcolor[HTML]{E7E6E6}0.0087               & \cellcolor[HTML]{E7E6E6}\textless{}0.0001    \\
\textbf{20}   & 0.8499               & 0.1198               & \textgreater{}0.9999 & \cellcolor[HTML]{E7E6E6}0.0009               & 0.0607               & \textgreater{}0.9999 \\
\textbf{40}   & \cellcolor[HTML]{E7E6E6}0.0009               & \textgreater{}0.9999 & 0.3003               & \cellcolor[HTML]{E7E6E6}0.0002               & 0.3972               & 0.1198               \\
\textbf{80}   & \cellcolor[HTML]{E7E6E6}\textless{}0.0001    & \cellcolor[HTML]{E7E6E6}0.0132               & \cellcolor[HTML]{E7E6E6}0.0132               & 0.2240                & 0.2240                & \textgreater{}0.9999 \\
\textbf{160}  & \cellcolor[HTML]{E7E6E6}\textless{}0.0001    & \cellcolor[HTML]{E7E6E6}0.0023               & 0.0197               & 0.5185               & 0.1198               & \textgreater{}0.9999 \\
\textbf{320}  & \cellcolor[HTML]{E7E6E6}\textless{}0.0001    & 0.1198               & \cellcolor[HTML]{E7E6E6}0.0003               & 0.0607               & \textgreater{}0.9999 & 0.5185               \\
\textbf{640}  & \cellcolor[HTML]{E7E6E6}\textless{}0.0001    & 0.1649               & \cellcolor[HTML]{E7E6E6}0.0005               & \cellcolor[HTML]{E7E6E6}0.0036               & 0.5185               & 0.5185               \\
\textbf{1280} & \cellcolor[HTML]{E7E6E6}\textless{}0.0001    & \cellcolor[HTML]{E7E6E6}0.0001               & \cellcolor[HTML]{E7E6E6}0.0009               & \textgreater{}0.9999 & \textgreater{}0.9999 & \textgreater{}0.9999\\
\hline
\multicolumn{7}{c}{\Precision}\\
\hline
\textbf{10}   & 0.2240               & \cellcolor[HTML]{E7E6E6}\textless{}0.0001 & \cellcolor[HTML]{E7E6E6}\textless{}0.0001    & \cellcolor[HTML]{E7E6E6}0.0002               & \cellcolor[HTML]{E7E6E6}0.0057               & \textgreater{}0.9999 \\
\textbf{20}   & \textgreater{}0.9999 & \cellcolor[HTML]{E7E6E6}\textless{}0.0001 & \cellcolor[HTML]{E7E6E6}0.0001               & \cellcolor[HTML]{E7E6E6}\textless{}0.0001    & \cellcolor[HTML]{E7E6E6}0.0009               & \textgreater{}0.9999 \\
\textbf{40}   & \textgreater{}0.9999 & \cellcolor[HTML]{E7E6E6}\textless{}0.0001 & \cellcolor[HTML]{E7E6E6}0.0197               & \cellcolor[HTML]{E7E6E6}\textless{}0.0001    & \cellcolor[HTML]{E7E6E6}0.0197               & 0.3003               \\
\textbf{80}   & \textgreater{}0.9999 & 0.5185            & \cellcolor[HTML]{E7E6E6}0.0036               & \textgreater{}0.9999 & \cellcolor[HTML]{E7E6E6}0.0423               & 0.5185               \\
\textbf{160}  & \textgreater{}0.9999 & 0.5185            & \textgreater{}0.9999 & \textgreater{}0.9999 & \textgreater{}0.9999 & \textgreater{}0.9999 \\
\textbf{320}  & \textgreater{}0.9999 & 0.3003            & \cellcolor[HTML]{E7E6E6}0.0023               & 0.6681               & \cellcolor[HTML]{E7E6E6}0.0087               & 0.6681               \\
\textbf{640}  & \textgreater{}0.9999 & 0.2240             & 0.2240                & \textgreater{}0.9999 & \textgreater{}0.9999 & \textgreater{}0.9999 \\
\textbf{1280} & \textgreater{}0.9999 & 0.4713            & 0.6141               & 0.1018               & 0.1423               & \textgreater{}0.9999\\
\hline
\multicolumn{7}{c}{\Recall}\\
\hline
\textbf{10}   & \cellcolor[HTML]{E7E6E6}0.0087            & \textgreater{}0.9999 & \textgreater{}0.9999 & 0.1198               & \cellcolor[HTML]{E7E6E6}0.0057               & \textgreater{}0.9999 \\
\textbf{20}   & 0.2240             & \textgreater{}0.9999 & 0.2240                & \textgreater{}0.9999 & \textgreater{}0.9999 & \textgreater{}0.9999 \\
\textbf{40}   & 0.8499            & \cellcolor[HTML]{E7E6E6}0.0132               & \cellcolor[HTML]{E7E6E6}0.0001               & 0.6681               & \cellcolor[HTML]{E7E6E6}0.0291               & \textgreater{}0.9999 \\
\textbf{80}   & \cellcolor[HTML]{E7E6E6}0.0003            & \cellcolor[HTML]{E7E6E6}0.0057               & \cellcolor[HTML]{E7E6E6}0.0197               & \textgreater{}0.9999 & \textgreater{}0.9999 & \textgreater{}0.9999 \\
\textbf{160}  & \cellcolor[HTML]{E7E6E6}\textless{}0.0001 & \cellcolor[HTML]{E7E6E6}0.0014               & 0.1649               & \textgreater{}0.9999 & 0.1649               & 0.8499               \\
\textbf{320}  & \cellcolor[HTML]{E7E6E6}\textless{}0.0001 & 0.1198               & \cellcolor[HTML]{E7E6E6}\textless{}0.0001    & 0.2240                & \textgreater{}0.9999 & 0.1649               \\
\textbf{640}  & \cellcolor[HTML]{E7E6E6}\textless{}0.0001 & 0.5185               & \cellcolor[HTML]{E7E6E6}0.0003               & \cellcolor[HTML]{E7E6E6}0.0057               & \textgreater{}0.9999 & 0.1198               \\
\textbf{1280} & \cellcolor[HTML]{E7E6E6}\textless{}0.0001 & \cellcolor[HTML]{E7E6E6}0.0023               & \cellcolor[HTML]{E7E6E6}0.0009               & \textgreater{}0.9999 & \textgreater{}0.9999 & \textgreater{}0.9999\\
\hline
\multicolumn{7}{c}{\Specificity}\\
\hline
\textbf{10}   & 0.2240                & \cellcolor[HTML]{E7E6E6}\textless{}0.0001 & \cellcolor[HTML]{E7E6E6}0.0001 & \cellcolor[HTML]{E7E6E6}0.0009               & 0.1649 & 0.6681               \\
\textbf{20}   & \textgreater{}0.9999 & \cellcolor[HTML]{E7E6E6}\textless{}0.0001 & \cellcolor[HTML]{E7E6E6}0.0001 & \cellcolor[HTML]{E7E6E6}0.0001               & \cellcolor[HTML]{E7E6E6}0.0014 & \textgreater{}0.9999 \\
\textbf{40}   & \textgreater{}0.9999 & \cellcolor[HTML]{E7E6E6}\textless{}0.0001 & \cellcolor[HTML]{E7E6E6}0.0423 & \cellcolor[HTML]{E7E6E6}\textless{}0.0001    & \cellcolor[HTML]{E7E6E6}0.0197 & \cellcolor[HTML]{E7E6E6}0.0423               \\
\textbf{80}   & \textgreater{}0.9999 & 0.1649            & \cellcolor[HTML]{E7E6E6}0.0002 & 0.5185               & \cellcolor[HTML]{E7E6E6}0.0014 & 0.3003               \\
\textbf{160}  & \textgreater{}0.9999 & \cellcolor[HTML]{E7E6E6}0.1198            & 0.1649 & 0.0858               & 0.1198 & \textgreater{}0.9999 \\
\textbf{320}  & \textgreater{}0.9999 & 0.3003            & \cellcolor[HTML]{E7E6E6}0.0014 & \textgreater{}0.9999 & \cellcolor[HTML]{E7E6E6}0.0197 & 0.5185               \\
\textbf{640}  & 0.5185               & \cellcolor[HTML]{E7E6E6}0.0057            & \cellcolor[HTML]{E7E6E6}0.0009 & 0.6681               & 0.2240  & \textgreater{}0.9999 \\
\textbf{1280} & 0.5185               & \cellcolor[HTML]{E7E6E6}0.0036            & \cellcolor[HTML]{E7E6E6}0.0014 & 0.5185               & 0.3003 & \textgreater{}0.9999\\
\hline
\end{tabular}
\end{table*}

\begin{figure*}[ht!]
\centering
\includegraphics[width=0.92\textwidth]{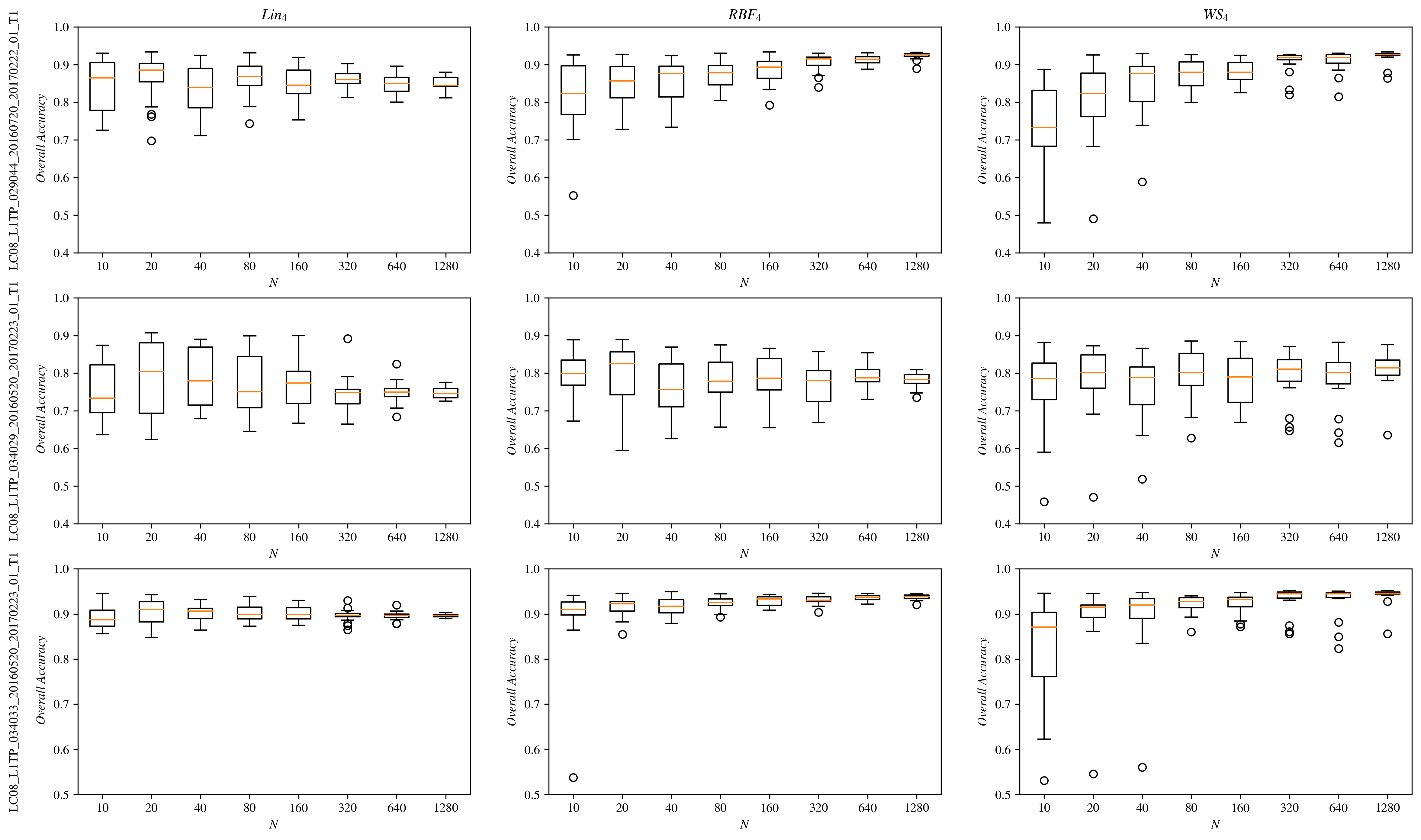}
\caption{Box plots showing \Accuracy\,obtained within 20 independent executions (for 20 reduced training sets of the $N$ size) for the three scenes shown in Figs.~\ref{fig:LC08_L1TP_034033_20160520_20170223_01_T1}--\ref{fig:LC08_L1TP_029044_20160720_20170222_01_T1}.}
\label{fig:Boxplot}
\end{figure*}

We are aware of some limitations of the hybrid SVMs. In Fig.~\ref{fig:Boxplot}, we render the box plots obtained for three test scenes visualized in Figs.~\ref{fig:LC08_L1TP_034033_20160520_20170223_01_T1}--\ref{fig:LC08_L1TP_029044_20160720_20170222_01_T1}. Although the aggregated metrics, averaged across 20 independent executions indicate that the $WS_4$ model is competitive with the classical RBF SVMs, the former classifier is slightly less stable, especially for lower $N$'s. However, increasing the size of the refined training set not only does allow for significantly enhance the generalization capabilities of the quantum-kernel SVMs, but it also improves their training stability. The best results (overall accuracy of approximately $92$--$93\%$) of the proposed simple machine learning models based on superpixel segmentation and SVMs do not deviate to a large extent from the current state-of-the-art deep learning models benefiting from the fully-convolutional architectures (overall accuracy of approximately $94$--$96\%$ reported for the 38-Cloud test scenes~\cite{38-cloud-1, 38-cloud-2, zhu2015improvement}). Such large-capacity deep learning models, however, can effectively exploit the contextual information within the image during the segmentation process---this may be of paramount importance for cloud detection, as the objects of interest may manifest different shape and spectral characteristics. Thus, designing additional feature extractors~\cite{Mahajan2020}, followed by feature selectors~\cite{Fei2022}, may be pivotal to further improve the classification accuracy of hybrid SVMs---appropriate feature extraction and fusion strategies have been shown extremely important in satellite image analysis using machine learning techniques~\cite{Chang2021JSTARS}.

\section{Conclusions and Future Work}\label{sec:conclusions}

In this work, we introduced hybrid SVMs exploiting quantum kernels for the task of cloud detection in multispectral satellite images, which is the ``hello, world'' in remote sensing. Such quantum-kernel models, together with classical SVMs with RBF and linear kernels were thoroughly investigated in the experimental study performed over a widely-used 38-Cloud dataset capturing Landsat-8 imagery. In our processing chain, the superpixel-powered training set selection is utilized to dramatically reduce the SVM training sets, and to pick the most informative training examples, together with the training prototypes, which are likely to become support vectors during the training process. Overall, we quantitatively, qualitatively, and statistically evaluated six SVM models---classical linear and RBF kernel-based SVMs, alongside the suggested hybrid SVMs based on the kernels elaborated by utilizing simulated quantum circuits called $S_4$, $WS_4$, $ES_4$, and $WSWS_4$. The hybrid model $S_4$ executed a \textit{stiff} (no variational layers) classical data encoding into separate qubits, $WS_4$ introduced one variational layer, $ES_4$ added the entanglement between the qubit registers, while $WSWS_4$ was a straightforward extension of the $WS_4$ model achieved by doubling it (two encoding layers interwoven with two variational layers).

The first observation inferred from our experiments is that the \textit{stiff} encoding $S_4$ model under-performs, when compared to the overall accuracy with other models (see the results rendered in Fig. \ref{fig:Acc_N}). Being able to embed data into vectors residing in 16-dimensional complex linear space does not necessarily increase expressivity and performance of the model---the linear kernel, defined on 4-dimensional space surpasses the $S$ model for all tested cases. Therefore, one needs to introduce additional parameters to the quantum feature map in order to control and tune its behavior. However, interestingly, there is no benefit in performance by introducing the 
entanglement via the layer $E$. 

The results reported here constitutes an exciting point of departure for further research. Albeit the classical and hybrid SVMs offer high-quality cloud detection, they are still slightly worse than the recent advancements in large-capacity deep learning models. This can be attributed to the fact that the SVMs investigated in this work operate on a small set of features that do not capture the subtle shape and spectral characteristics of the pixels' neighborhood. We anticipate that introducing new feature extractors to our pipeline can substantially enhance the classification capabilities of the models. Our research efforts are focused on deploying quantum-kernel SVMs for other multispectral data for cloud segmentation (and segmentation of other objects of interest as well, e.g., cultivated land~\cite{Tulczyjew2022GRSL}), especially in large-scale Sentinel-2 imagery, as well as on using them for hyperspectral image classification~\cite{Nalepa2019validating}, and on quantifying their robustness against noise-contaminated data~\cite{NalepaStanek2020IGARSS}. Finally, we are currently investigating the non-functional abilities of both classical and deep machine learning models, with a special emphasis put on their inference time, as it is critical in processing massively large amounts of satellite imagery captured nowadays.





%



\section*{Acknowledgment}

This work was funded by the European Space Agency,
and supported by the ESA $\Phi$-lab (https://philab.phi.esa.int/),
under ESA contract No. 4000137725/22/NL/GLC/my. 
JM, GC, and FS were supported by the Priority Research 
Area Digiworld under the program Excellence Initiative 
– Research University at the Jagiellonian University 
in Krak\'ow. JN was supported by the Silesian University of Technology grant for maintaining and developing research potential.

\ifCLASSOPTIONcaptionsoff
  \newpage
\fi



\bibliographystyle{IEEEtran}
\bibliography{bibliography}
\end{document}